# KPIs-Based Clustering and Visualization of HPC jobs: a Feature Reduction Approach

**Mohamed Soliman Halawa[1], Rebeca P. Díaz Redondo[2], Ana Fernández Vilas[2]**

[1]Business Information Systems Department, Arab Academy for Science Technology and Maritime Transport, Cairo, Egypt (e-mail: halawamohamed@aast.edu)

[2]Information & Computing Lab, atlanTTIC Research Center, Universidade de Vigo, Spain (e-mail: rebeca@det.uvigo.es; avilas@det.uvigo.es)

Corresponding author: Mohamed Soliman Halawa (e-mail: halawamohamed@aast.edu).

This work was supported by the European Regional Development Fund (ERDF) and the Galician Regional Government, under the agreement for funding the atlanTTIC Research Center for Information and Communication Technologies (atlanTTIC), and the Spanish Ministry of Economy and Competitiveness, under the National Science Program (TEC2017-84197-C4-2-R). The authors would also like to thank the Supercomputing Center of Galicia (CESGA) for their support and the resources for this research.

**ABSTRACT** High-Performance Computing (HPC) systems need to be constantly monitored to ensure their stability. The monitoring systems collect a tremendous amount of data about different parameters or Key Performance Indicators (KPIs), such as resource usage, IO waiting time, etc. A proper analysis of this data, usually stored as time series, can provide insight in choosing the right management strategies as well as the early detection of issues. In this paper, we introduce a methodology to cluster HPC jobs according to their KPI indicators. Our approach reduces the inherent high dimensionality of the collected data by applying two techniques to the time series: literature-based and variance-based feature extraction. We also define a procedure to visualize the obtained clusters by combining the two previous approaches and the Principal Component Analysis (PCA). Finally, we have validated our contributions on a real data set to conclude that those KPIs related to CPU usage provide the best cohesion and separation for clustering analysis and the good results of our visualization methodology.

## I. INTRODUCTION

Research in different areas is currently highly dependent on intensive computation tasks and/or complex simulations [1]. High-Performance Computer (HPC) systems provide the computing infrastructure needed to carry out these tasks. However, HPC systems are inherently complex and costly systems that need specific monitoring systems to ensure their reliability [2]. These monitoring systems are constantly checking the performance of each one of the vast number of nodes [3] by collecting the information obtained by sensors. These performance indicators, also known as Key Performance Indicators (KPIs) [4] are usually grouped under different categories, like CPU usage, memory usage, network traffic, and other hardware sensors. The information received is usually stored as time series: each value consists of the reading (KPI value) and the time (date and time) when it was collected. Thus, abnormal variations in KPI time series may be an early evidence of a problem in the execution of a job and/or in the usual behavior of a node in the HPC system [5]. The early detection of these anomalies is critical to help the system administrators take proactive measures to identify the source of the problem and to prevent more serious escalation of the problem, which may cause job failures.

However, the HPC system generates a tremendous amount of KPIs that are very hard to analyze. The KPIs are collected via sensors from jobs executing on hundreds of parallel nodes daily, with a high frequency (approximately one sample each minute), and are usually stored as time series. Thus, analyzing these time series is a very challenging and computationally costly problem. Additionally, and since this data is not labeled, some approaches, as clustering, are considered the most adequate to face the analysis of the KPI time series [6], [7]. Time series clustering helps to identify patterns of performance behaviors by grouping similar time series. However, time series clustering is a complex problem due to its high data dimensionality, which may cause problems such as highly biased estimates and decreases clustering performance [6]. Therefore, dimensionality reduction techniques play an important role in this field [8] by transforming a high-dimensional data representation into a lower low-dimensional data representation [9].

In this study, we propose a new methodology to cluster and visualize HPC jobs based on their performance KPIs. In order to deal with the high dimensionality problem (inherent to time series and aggravated by the huge amount of collected data), we propose to face the time series clustering in terms of their features, usually related to statistical behavior or global parameters (trend, seasonality, skewness, periodicity, etc.) of the time series. The underlying idea is selecting only those features that are relevant for the job clustering, assuming that these features are usually correlated and redundant. With this aim, we propose a twofold feature selection: on the one hand, a literature-based feature selection, and, on the other hand, a variance-based feature selection. We also compare the obtained results to the results of a previous approach [7] that tried to face HPC jobs clustering by applying Principal Component Analysis (PCA) techniques. Our second contribution, additionally to the time series clustering proposal, deals with the visualization problem of clustering for massive data. We propose applying the same mechanism of feature selection used for clustering to focus the visualization tasks only on the two and three most relevant features, to create a 2D and 3D plot, respectively.

In order to validate our contributions (clustering and visualization), we have created a data set with data gathered from the Centro de Supercomputación de Galicia (CESGA): 195 running nodes, 11 KPIs, 9,006 jobs executed in a period of ten months and a sample rate from 60 to 120 seconds between samples. These characteristics entailed a data set with 35,761,300 samples per KPI and per job. Our results are promising since visualization is already much more effective, and the quality clustering metrics offer better results than in our previous analysis [7].

To the best of our knowledge, the main novelties of our analysis are the following ones: (i) the wide variety of extracted features used for the job clustering: KPIs related to CPU usage, memory usage, network traffic, and other hardware sensors; (ii) the twofold feature selection techniques used to identify the most relevant features; (iii) the methodology to help in the visualization tasks and (iv) the selection of the two-dimensionality reduction techniques features applied in our previous work [7] and the feature extraction applied in this study to cluster the HPC jobs.

The rest of the paper is organized as follows. Section II summarizes the background about the techniques used in this study, whereas Section III offers an overview of the most relevant approaches in the state-of-the-art in the time series dimensionality reduction field and in the HPC analysis field. The data set used for our analysis is described in Section IV. Section V describes the methodology for clustering and visualization. Results are summarized in Section VI. Finally, discussion is in Section VII, and Section VIII covers the conclusions and future work.

## II. BACKGROUND

The emergence of ubiquitous sensing infrastructure offers the possibility of gathering huge volume of data from numerous sensors, which usually send this information frequently [10], [11]. Within this context, time series data offer an appropriate mathematical framework to gather and analyze the collected information. However, time series data inherently entails a high dimensionality of the data. Each time series length is high and usually much larger than the number of time series (number of sensors). Additionally, the number of features that can be extracted from each time series is also high. Considering both (i) the large number of time series (number of sensors) and (ii) their extracted features involvement in any analysis causes the so-called Curse of the Dimensionality [12], which also entails problems with calculation of distance metrics. In this section, we summarize previous approaches in two supplementary fields related to our proposal: (i) time series dimensionality reduction, as a way to face the high dimensionality problem; and (ii) clustering techniques, since we propose to use this mathematical technique for our research work.

### A. TIME SERIES DIMENSIONALITY REDUCTION

Analyzing time series with the aim of checking similarities usually involves a non-trivial problem due to the high dimensionality of the data. In the specialized literature, there are several approaches that face the dimensionality reduction issue from different perspectives.

Some techniques are based on representing the data under a different form, which simplifies the data volume, in such a way that minimizes the global reconstruction error. Within this field, there are some remarkable approaches, like Singular Value Decomposition (SVD) [13], the Discrete Fourier transform (DFT) [14] or the Discrete Wavelet Transform (DWT) [15]. Other proposals are based on a different philosophy: to obtain a new and approximate time series that summarizes the behavior of the original one. Within this field, there are some interesting approaches like the Piecewise Aggregate Approximation (PAA), which divides the original time series into sections and records mean values of these sections for analysis, and its evolution into a more dynamic technique[16], coined as Adaptive Piecewise Constant Approximation (APCA) [17].

Other approaches do not face a transformation of the time series directly, but a selection or extraction of relevant features to be analyzed. These lines of research, also known as dimensionality reduction, are based on the following assumption: depending on the nature of the time series, its features are usually correlated and redundant. Usually, approaches in this line are categorized into two groups [6], [18], [19]: feature selection and feature extraction. The former focuses on selecting the most significant features from the original data set by applying different techniques, such as univariate statistical tests, variance thresholds, or Principal Component Analysis (PCA) [20], [21], [22], [23].

The selection of the best feature set only depends on the data set and there is not a selection technique that performs well with any data set [24]. The variance threshold technique [25] is considered the simplest and most commonly used; it selects as relevant features those of them whose variance is greater than a threshold:

$$J_{Variance}(X_k) = \frac{1}{n-1}\left(X_i^{(k)} - \frac{1}{n}\sum_{i=1}^{n} X_i^{(k)}\right)^2 \quad (1)$$

Where K is the identifier of each time series, n is the number of samples in the time series, and $x_i^{(k)}$ is the iteration instance of the time series $X_k$, i ∈ {1, …, n}.

Feature extraction, on another hand, is based on extracting new features from the original data set. In the time series field, these new features are usually related to statistical behavior or global parameters of the time series [26], like trend, seasonality, periodicity, skewness, etc. Thus, the underlying idea is obtaining a time series summary in terms of its correlating structure, distribution, entropy, etc. [27], which gives interesting information about the time series characteristics and enables different analysis (classification, clustering, comparison, outlier detection, etc.). Within this line, a time series feature extraction package in Python was recently published, called Time Series FeatuRe Extraction on the basis of Scalable Hypothesis tests (Tsfresh) [28], which is widely used because of its clear advantages. First, it was conceived precisely to identify and extract meaningful features from time series, so it is able to combine 63 time series characterization methods to compute (by default) a total of 794 features. Second, it implements standard APIs, so it can be used together with other machine learning libraries (e.g. Scikit-learn [29], Numpy [30], or Pandas [31]). Finally, it also includes methods to evaluate the power and importance of these characteristics for regression or classification purposes.

### B. CLUSTERING ALGORITHMS

Clustering algorithms try to split up a set of data objects into subsets or clusters that group objects similar to one another, but dissimilar to objects in other clusters. There are multiple clustering methods, which are generally classified in the following four categories [32]: (i) Partitioning methods: Data points are organized in a given number $k$ of groups, so there is at least an element in each group; (ii) Grid-based methods: The object space is quantized into a finite number of cells that form a grid structure on which the clustering is performed; (iii) Hierarchical methods: The data set is decomposed into multiple levels, organizing the data into a tree of clusters. It is quite inflexible, since once a step is done it cannot be undone; and (iv) Density-based clustering methods: The notion of density is used. The general idea is to grow a given cluster while its density excesses a given threshold. This way clusters are formed in dense areas while the points in sparse areas are considered as noise.

Cluster analysis has been the solution in many application domains like text mining, information retrieval, social network analysis, image and video processing, bioinformatics, image processing, and so on in past years [33], [34]. Clustering algorithms are used to identify homogeneous groups of objects and useful patterns in a dataset [35], [36]. Cluster analysis offers a diverse number of clustering algorithms such as K-means, Db-scan, Hierarchical clustering, and Gaussian Mixture Model (GMM) [37].

One of the most widely used clustering techniques is K-means [34], [37], [38] a partition method that organizes data into K-clusters. K is predefined, by assigning each object (data observation) to the cluster with the nearest centroid, which is usually measured by using the Euclidean distance [39]. The algorithm begins defining K centroids (randomly) and then uses multiple iterations to update these centroids in such a way that the distance values of data objects and the nearest centroids are minimized. The iterations end when the ideal centroids are identified:

$$d(\boldsymbol{p}, \boldsymbol{q}) = \sqrt{\sum_{i=1}^{n}(p_i - q_i)^2} \quad (2)$$

where $d(\boldsymbol{p}, \boldsymbol{q})$ represents the distance between two n-dimensional vectors $\boldsymbol{p}$ and $\boldsymbol{q}$.

Thus, K-means cluster centroid is considered the arithmetic mean of all the objects inside this cluster [40]. K-means was successfully applied in different application fields, from image processing to parallelization or to analyze behavior in urban areas [33], [35], [36], [40].

K-shape [41] was designed specifically for time series clustering. This method relies on a scalable and iterative refinement procedure that assesses the distance among clusters by using a normalized version of the cross-correlation measure, which is used to take into account the shapes of time series in the comparison process. Thus, contrary to other approaches, K-shape considers the shapes instead of treating the observations in time series as independent attributes. Based on the properties of the shape-based distance measure, the algorithm computes the centroids, used to capture the share characteristics of the data and to assign time series to clusters. Its robustness has been experimentally evaluated against other partition methods, like K-means or K-medoid [42]. K-means was chosen for this study due to its simplicity and efficiency with large datasets. K-means usually offers faster computation results than other algorithms when K is small. Furthermore, the clusters that results from the K-means algorithm are usually more cohesive than other algorithms such as hierarchical clustering [43]. Lastly and most importantly, the visualization methodology applied in our proposed framework (Section V.C) depends on the K-means clustering centroids, which were used to rank features influences to the clustering model to plot the clustering results.

Most of clustering algorithms consider all feature components of the data to be equally important. However, some of them are irrelevant and might cause incorrect clustering results. This is especially relevant when talking about time series, since the high number of features is one of the issues when applying clustering. Consequently, some approaches, usually known as Feature-weighted techniques, have arisen to take into account the importance of features before performing the clustering algorithms. This is the case of the Simultaneous Weighting on Views and Features (SWVF) [44] or the Weighted Multi-view Clustering with Feature Selection (WMCFS) [45]. Both of them assign weights to features before clustering, however they do not reduce their number. This approach was adopted by Feature-Reduction Multi-View K-means (FRMVK) [46], where irrelevant features are directly removed from the analysis.

This is, precisely, the approach we have followed in this case, applied to a different context but with the same underlying philosophy: trying to detect those features in the time series data that are not relevant for the analysis and that might be removed before applying the clustering algorithm.

### C. CLUSTER VALIDATION TECHNIQUES

Quality assessment of the clustering results is key to validate the analysis [47] and might help to identify, for instance, the best optimal number of clusters in a data set [48], beyond the experience of the data analysts. Some validation methods require from labeled data to evaluate the goodness of the clusters (external methods [49]), whereas others do not need from this information (internal methods). In our research work, and since we do not have tagged data, we apply internal methods. Usually, clustering validation techniques are based on two criteria. On the one hand, compactness, i.e. the members of each cluster should be as close to each other as possible. Variance is a common measure of compactness, which must be minimized within each cluster. On the other hand, separation, i.e. the clusters should be widely spaced. There are three common approaches to measure the distance between two clusters: (i) single linkage, which measures the distance between the closest members of the clusters; (ii) complete linkage, which measures the distance between the most distant members; and (iii) comparison of centroids, which measures the distance between the centers of the clusters.

There are three popular approaches that take into account the two previously mentioned criteria (compactness and separation): (i) the Silhouette coefficient [50]; (ii) the Calinski-Harabasz index [49] and (iii) the Davies-Bouldin index [51]. However, the Silhouette index (SH) is, because of its simplicity, the most widely used. It generates a score between -1 and 1, which is independent from the clustering algorithm, and that represents the quality of the cluster results [49]. A SH score can be interpreted as follows: a score between 0.71 and 1 shows excellent clusters results, a score between 0.51 and 0.70 shows acceptable clusters results, a score between 0.26 and 0.50 shows poor clusters results, and, finally, less than 0.25 is considered a not acceptable result [52] [53].

$$SH(i) = \frac{b(i) - a(i)}{\max\{b(i); a(i)\}} \quad (3)$$

where $a(i)$ represents the mean distance between an object $i$ and each point within the same cluster and $b(i)$ represents the mean distance between an object $i$ and the points in all the other clusters.

### III. RELATED WORK

As it was previously mentioned, feature extraction is based on extracting new features from the original data and, in the time series field, these new features are usually related to the statistical behavior or the global parameters of the time series. This approach was previously applied in different application fields with the aim of reducing the number of variables (features) for the analysis. After an exhaustive analysis of the specialized literature, Table 1 summarizes previous approaches that have used different time series features for their analysis, such as skewness, mean, trend, variance, seasonality, etc.

There are interesting proposals in this line of research applied to different areas. For instance, the authors of [54] and [55] tried to discover the best forecasting method based on the accuracy of the predictions, using extracted measures such as trend, seasonality, periodicity, etc. Within the classification field, [56] proposes a hybrid algorithm (Zeus) to classify lightning time series data. This algorithm uses progressive computation for features extraction and selection, which were fed for classification using a support vector machine (SVM) [57]. It reduces, to a great extent, the number of features needed. Also, in [58] a new feature-based approach is proposed to recognize patterns in drilling time series data. In this case, the drilling data dimensionality was reduced using feature extraction and selection methods to improve the accuracy of the classifiers. The approach improved the rate of classification accuracy by 10% with a faster learning time. Feature extraction was also successfully used for anomaly detection in different areas. For instance, the authors of [59] and [60] proposed a framework coined as EGADS to identify anomalous time series by using some relevant features (trend, frequency, seasonality, etc.). This approach achieved more accurate and faster detection than other compered methods.

We have also found some proposals that focus on clustering time series by reducing the number of features to be analyzed. This is the case of the work in [61] where clustering of time series is based on their structural characteristics instead of clustering point values using a distance metric. The set of features used (trend, seasonality, periodicity, serial correlation, skewness, kurtosis, chaos, nonlinearity, and self-similarity) achieved high accuracy clusters with benchmark time series datasets. In [62] the research work focused on farming, clustering farms based on similar time series statistical features of meat inspection in pig slaughterhouses. This information was useful to detect farm level risk factors causing health problems based on disease incidence and trend.

Going deeper in the use of this technique in the analysis of HPC systems, there are very interesting approaches that extract features from the time series data in order to perform a more efficient analysis. Geelen et al. [63], for instance, presented a method for detecting and localizing pipe leakage using real-time pressure sensor measurements named Monitoring Support. The method applied both time series instance-based and feature-based clustering to detect recurring pressure anomalies, which are suspected to be damages or leaks of the water distribution network. The results showed that feature-based clustering performed better in detecting anomalies on two pressure sensor data sets with accuracy F1-scores of 92% and 94%. Klinkenberg et al. [64] proposed a supervised prediction model to predict nodes' failures in HPC systems. The prediction model used extracted statistical features of the time series frame from the nodes monitoring data before a failure occured. This prediction model was able to classify defected nodes with a recall of 91% and a precision of 98%. Tuncer et al. [65] proposed a new framework to automatically detect previously occurred performance anomalies in HPC systems using applications performance counter data. In the framework, statistical features were extracted from the application's performance counters time series data to reduce the data scale and improve the performance of the machine learning algorithms to identify the anomalies. The evaluation of the proposed framework showed outstanding results in detecting anomalies with F-score over 0.97. Tuncer et al. [66] proposed a framework to classify performance variations in HPC systems. The framework extracted several statistical features of the KPIs collected by the HPC monitoring system. After, these features were used to classify the system behavior using different machine learning algorithms. The results showed that the Random Forest algorithm achieved the best accuracy. Frank et al. [67] tried to identify failed nodes that are being used by running large-scale applications on the HPC system. The authors proposed a new feature-based system for node failure predictors using machine learning with a low percentage of false alarms at large scales. The failure prediction system proved its usefulness after achieving a reused unnecessarily triggering checkpoints (UC) with a lead-up time of four minutes in a large production cluster. Finally, Jin et al. [68] designed a feature-categorizing-based hybrid system to detect anomalies in communication systems. The proposed system extracted statistical features from each KPI time series. Then, the extracted features were categorized by the KPI category identifier to different groups of features based on similar statistical characteristics. Afterwards, each feature group was fed to the applicable anomaly detector technique. Finally, aggregation of the results of all the anomaly detector techniques was used to detect an anomaly in terms of the entire feature space. The system has proven it could detect different anomalies with lower false alarms.

**Table 1.** Time series features

| Features | References | Features | References |
|---|---|---|---|
| Skewness | [54], [55], [56], [58], [61], [62], [63], [64], [65], [66] | Count above mean | [60] |
| Kurtosis | [54], [55], [56], [58], [61], [62], [63], [64], [65], [66] | Count below mean | [60] |
| Mean | [56], [58], [59], [60], [62], [64], [66], [67], [68], | Historical change | [60] |
| Autocorrelation or Serial correlation | [54], [55], [59], [61], [62], [63], [65] | Simple moving average | [60] |
| Standard deviation | [55], [56], [58], [62], [63], [64], [67] | Weighted moving average | [60] |
| C3 (nonlinearity) | [54], [55], [61], [62], [63] | Percentiles | [58] |
| Max | [56], [58], [60], [63], [64], [66], [67], | Difference | [60] |
| Min | [56], [58], [60], [63], [64], [67], [66] | Integration | [60] |
| Trend | [54], [55], [59], [61], [62], | Friedrich_coefficients | [63] |
| Variance | [56], [58], [59], [64], [66], [67], | Linear_trend slope | [63] |
| Median | [58], [64], [67], [68] | Linear_trend intercept | [63] |
| Seasonality | [54], [55], [61], [62] | Large_standard_deviation | [63] |
| Periodicity (frequency) | [54], [61] | Remainder | [62] |
| Lyapunov exponent | [55], [62] | Length | [64] |
| Self-similarity | [54], [61], [65] | Integral | [56] |
| Season Strength of seasonality | [54], [59] | Step changes | [55] |
| Sum | [56], [58], [67] | Ratio of Means | [56] |
| Mode | [58], [68] | Number of peaks | [55] |
| Chaotic | [54], [61] | Peak | [59] |
| Spkt_welch_density(Powers pectrum) | [55], [63] | Trough | [59] |
| Hurst | [62] | Mean change | [60] |
| Entropy, spectral entropy | [58], [59] | Lumpiness | [59] |
| Partial_autocorrelation | [55], [63] | Predictability | [55] |
| Interquartile range | [58], [66] | Lshift Level shift using rolling window. | [59] |
| Range | [58] | Variance change. | [59] |
| Linearity | [59], [65] | Flat spots | [59] |

| Exponentially Weighted Moving Average | [60] | Crossing points | [59] |
|---|---|---|---|
| Mean second derivative central | [60] | Absolute sum of changes | [60] |
| Durbin–Watson statistic of regression residuals | [55] | Klscore | [59] |
| Time reversal asymmetry statistic | [63] | Spikiness | [59] |

Our proposal faces the KPIs analysis from a similar perspective to other previous approaches in the literature [64]-[68]. However, our approach has the following differences that allow us to state that we provide a wider analysis: (i) we gather data from different KPIs categories (CPU usage, memory usage, network traffic, and other hardware sensors); (ii) we have obtained a more representative data set, since the data collection was done for a long period of time and data was gathered from a large number of jobs in the HPC system; (iii) we have faced a pre-assessment of the different extracted features from time series in order to analyze the most representative ones by using two different approaches (variance-based and literature-based). Finally, we have also added a final step to visualize the job clusters based on ranking job features.

## IV. DATASET DESCRIPTION

HPC system includes hundreds of computational nodes that execute an enormous amount of jobs daily. The constantly increasing complexity of these systems also requires complex monitoring systems that obtain raw data about the usage of the available resources, such as disk, CPU, traffic, etc. These monitoring systems gather raw data or KPIs from each single computational node, which is organized into time series (one per metric), as Figure 1 shows.

For our analysis, we have obtained our data set from the CESGA Foundation (https://www.cesga.es/en). This Galician data center is a non-profit organization that has the mission to contribute to the advancement of Science and Technical Knowledge by means of research and application of high-performance computing and communications, as well as other information technologies resources. CESGA Foundation is considered an instrument for sustainable socio-economic development, devoting special attention to the relations of cooperation between research centers, whether public or private, and the productive sector. This is the framework that enables the cooperation between this institution and the Universidade de Vigo.

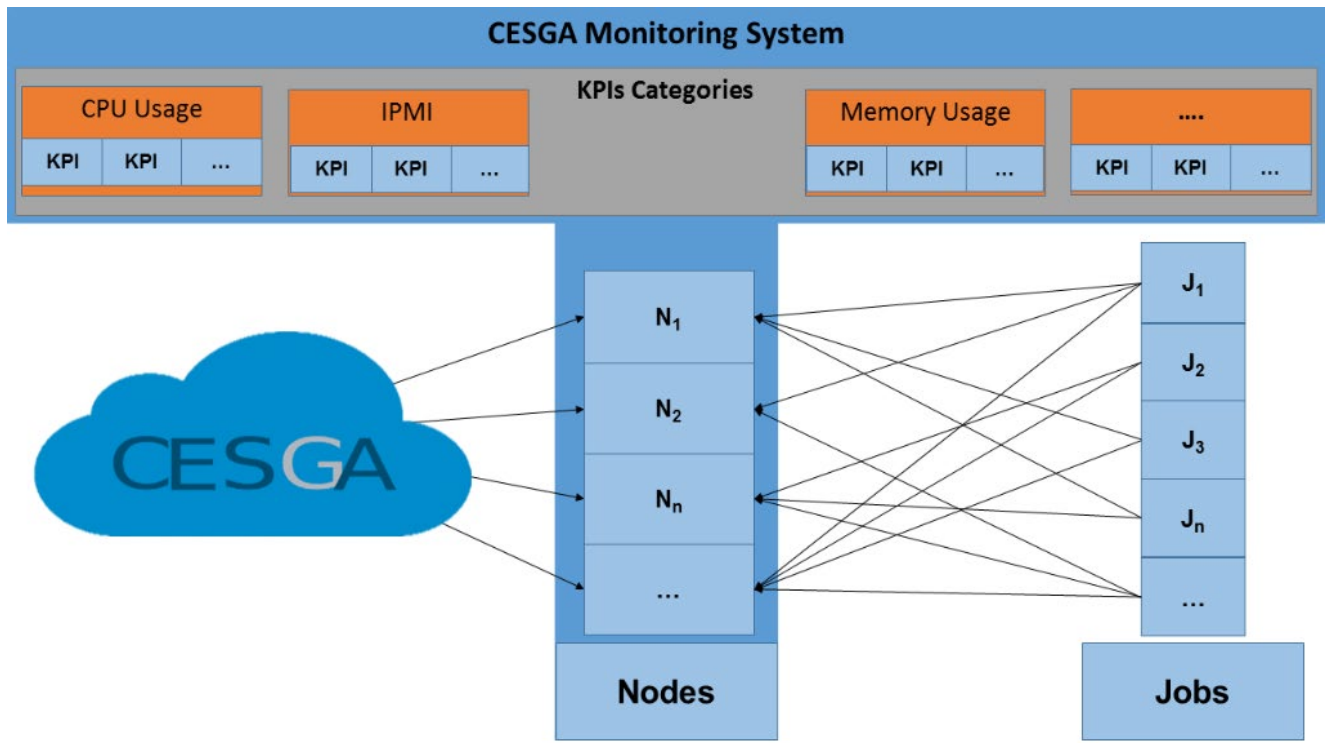

**FIGURE 1.** Cloud monitoring system.

**Table 2.** Selection of the performance metrics.

| Category | Metric | Definition |
|---|---|---|
| CPU usage | aggregation.cpu-average.percent.idle (IDLE) | The aggregated average Percent of time when the CPU is idle. |
| | aggregation.cpu-average.percent.system (SYSTEM) | The aggregated average Percent of time when the CPU is working. |
| | aggregation.cpu-average.percent.wait (WAIT) | The aggregated average Percent of time when the CPU is waiting |
| Network (interface) traffic | interface.bond0.if_octets.rx (RX) | The number of bytes received over the network per second. |
| | interface.bond0.if_octets.tx (TX) | The number of bytes transmitted over the network per second. |
| IPMI | IPMI.CPU1_Temp (CPU1) | The temperature readings of CPU1 |
| | IPMI.CPU2_Temp (CPU2) | The temperature readings of CPU2 |
| | IPMI.PW_consumption (PW) | The power consumed by the system hardware |
| | IPMI.System_Temp (SYSTEM_TEMP) | The temperature readings of the system |
| System Load | load.load.shortterm (SHORTTERM) | System load average over the last minute |
| Memory usage | memory.cached.memory (MEMORY) | Cached Memory occupied |

The CESGA center has different computing platforms with different available architectures to help users to select the best one for their computation requirements: Supercomputer Finisterrae II, Cloud Computing architecture, and a Big Data architecture [69]. First, Finisterrae II is a supercomputing platform that is usually used for highly intensive simulations and calculations. The Finisterrae II cluster is composed of 306 nodes, each one with two Haswell 2680v3 processors (24 cores) and 128 GB of RAM [69]. Second, the cloud computing platform provides a virtual and elastic computing infrastructure that can be customized to the end-user requirements (operating system, number of processors, etc.). Finally, the Big Data architecture supports the parallel processing of huge volumes of information by using the power of the state-of-the-art technologies and software focused on data management and processing. The Big Data platform is composed of a dedicated cluster of 38 nodes, each one with 2 Intel Xeon E5-2620 v3 processors (12 cores) and 64GB of RAM. These three computing platforms can be remotely accessed through a Secure Shell (SSH) terminal and a simple Web User Interface (WebUI) [69].

In this study, we have used the CESGA Big Data platform to extract our KPIs data. This platform uses the third version of the Hadoop framework and it offers the Hadoop ecosystem core components. Hadoop is an open-source framework that processes offline big data in a distributed manner. Hadoop is characterized for having different key components for this computation, such as the Hadoop Distributed File System (HDFS) [70], a MapReduce framework [71], the Hive warehouse [72], the Apache HBase distributed database [73], etc. HDFS is the primary storage in Hadoop and it manages the storage of data in blocks to be processed by any job [74]. Hadoop uses the MapReduce framework to perform the computation in two different stages: (i) the Map stage, which divides the task (computation) of the job in parts to be computed by different nodes (in parallel) and (ii) the Reduce stage, which processes and summarizes the data coming from the Map stage (different nodes) [74]. Hive is a Data warehouse infrastructure that uses SQL queries (HiveQL) to query data from Hadoop [72]. It converts the input program written in HiveQL into MapReduce jobs executed on Hadoop [75]. Finally, HBase is an open-source, column-based database built on top of the Hadoop file system, which offers random real-time read/write access to data in the Hadoop File System [70]. The CESGA Big Data platform uses Apache Spark [76] to improve the parallelism processes. Apache Spark is a unified analytics engine and a set of modules for parallel data processing on computer clusters [77]. It runs over Hadoop to speed the parallel processes since it may run on the memory and not on the disk, as Hadoop does [78].

**Table 3.** Performance metrics description.

| Category | Metric | Sample frequency | The metric length (Number of samples for all jobs) |
|---|---|---|---|
| CPU usage | IDLE | 60/120s | 35,761,301 |
| | SYSTEM | 60/120s | 35,761,297 |
| | WAIT | 60/120s | 35,761,300 |
| Network (interface) traffic | RX | 90-93s | 35,761,301 |
| | TX | 90-93s | 35,761,298 |
| IPMI | CPU1 | 90-93s | 35,761,301 |
| | CPU2 | 90-93s | 35,761,300 |
| | PW | 90-93s | 35,761,301 |
| | SYSTEM_TEMP | 90-93s | 35,761,299 |
| System Load | SHORTTERM | 60/120s | 35,761,301 |
| Memory usage | MEMORY | 60/120s | 35,761,200 |

FIGURE 2. Example of information gathered per KPI (n jobs and m nodes)

The HPC monitoring systems used in CESGA generate over 44,280 different KPIs from different categories, which is actually an overwhelming number of KPIs for any analysis. Therefore, and within the line of our previous collaboration with CESGA [7], we have followed the technical recommendation of the CESGA technicians to focus our analysis only on the 11 KPIs summarized in Table 2. These KPIs belong to five categories (CPU usage, memory usage, system load, IPMI, and network interface) and were selected according to their importance and clear representation of the performance of jobs.

The data from the 11 KPIs was collected by submitting a specific spark job through one of the Hadoop 3 cluster login nodes. The job was designed to gather the KPIs data using Hive queries from the Hadoop cluster (typically one master node and several slaves). The results of each KPI query was held in a dataframe using the Spark dataframe API. Finally, these dataframes are stored in the Hadoop Distributed File System (HDFS), just prepared for the analysis. We have chosen HDFS since it is the primary storage in Hadoop. The data extractor ran from February 1st, 2019 to October 31st, 2019 gathering data (as dataframes) from the HPC monitoring system of 195 parallel nodes, where a total number of 9,006 jobs were running. Each one of the 11 extracted KPI time series dataframe contains four columns with the following information: (i) the value of the KPI, (ii) the time of the machine when the value was acquired, (iii) the job, and (iv) the node to which this value belongs. Afterwards, for each KPI, the job information was filtered by its identifier to collect the time series information to each one of the nodes where the job was executed, as Figure 2 shows.

Each KPI has a different sampling frequency that variates from 60-120 seconds in the case of CPU usage metrics, System load, and Memory usage in Table 3 to the 90 seconds in the case of Network interface and IPMI system in Table 3. That means that, on average, we obtain 35,761,290 samples per KPI and per job: an overwhelming amount of data.

## V. METHODOLOGY

As it was previously mentioned, the challenge of analyzing data gathered by the HPC monitoring systems is the huge amount of information that is collected. Each computing node is monitored by a set of sensors that are sampling the information of a high number of variables or KPIs (CPU usage, temperature, humidity, memory usage, etc.) and store this data as a set of time series, one per KPI. Each job, which is running on a set of nodes, has as many data about its behavior as the KPIs from all the nodes where it is executed. In our case, and as it was summarized in the previous section, we have collected information from 11 KPIs that monitor 195 nodes where 9,006 jobs were executed; on average, we have gathered 35,761,300 samples per KPI and per job.

We propose to face this massive analysis by reducing the large scale and dimensionality as it is summarized in Figure 3. With this aim, we firstly (phase 1 in Figure 3) extract a set of features per each one of the time series (or KPI). These features represent the statistical behavior and general parameters of the KPI (time series): such as skewness, mean, trend, variance, seasonality, etc. In our proposal, we apply the Phyton extraction package Tsfresh [28], which has several advantages (previously mentioned in Section II.B). Section V.A details the different steps within this first phase.

Then (phase 2 in Figure 3), we propose to select only those features that are, indeed, relevant and perform a clustering analysis. For the first sub-task (step 2.1: feature selection) we trust in two different and supplementary approaches. On the one hand, we consider the information collected from other approaches in the specialized literature to analyze each KPI (time series) individually. On the other hand, we take into account the variance threshold using three different percentages (80%, 85%, and 90%). For the second sub-task (step 2.2: clustering) we conducted two experiments using the K-means algorithm (Section V.D): one of them analyzing the KPIs (times series) individually and the other using a combined approach. In both cases, we used the Silhouette index to determine the optimal number of clusters. Section V.B details the different steps within this second phase.

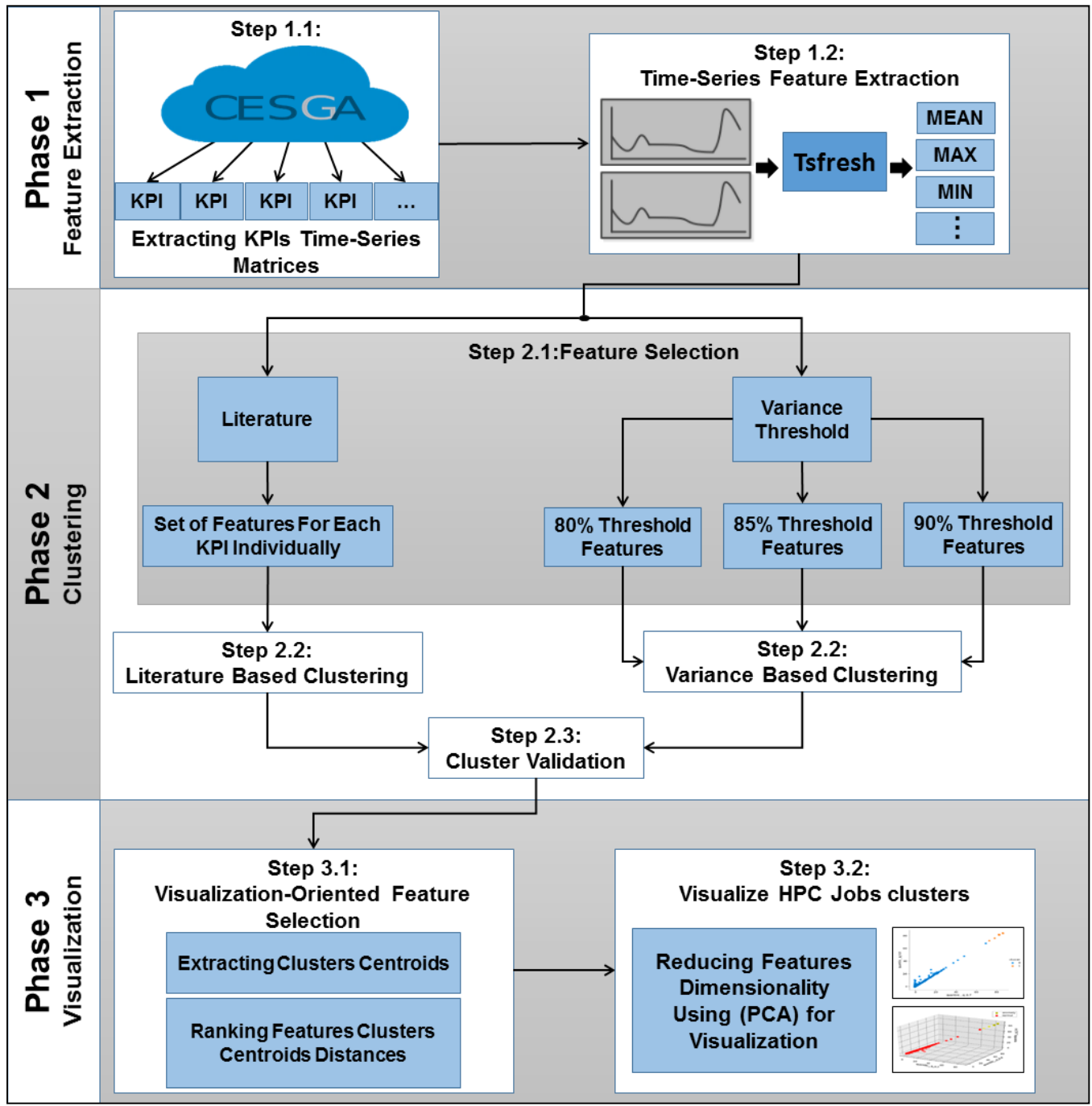

**FIGURE 3.** HPC job clustering and visualizing methodology using time series feature extraction.

Finally (phase 3 in Figure 3), we use the obtained clusters to visualize the data (Section V.C). With this aim, we calculated the cluster centroids and, using the Euclidean distance, we tried to identify the three features that have the most influence on the K-means clustering. These three features were used to provide 2D and 3D plots that help to visualize the clustered jobs and to identify the common characteristics.

### A. PHASE 1: TIME SERIES FEATURE EXTRACTION

As it was previously mentioned, we have opted to use time series features to represent their characteristics and behavior in an attempt to reduce the amount of data and the dimensionality problem consequence of a huge data set. With this aim, we have used the Python library Tsfresh [28]. Although this library supports the extraction of 794 features per time series, we have decided to extract the 36 features summarized in Table 4, whose definitions are stated in Table A (appendix) [4].

This first selection was done taking into account the usual features used in the literature, as well as our own experience in time series analysis. The extracted features were labeled as follows: "NodeID_KPIName_FeatureName", where KPIName is taken from Table 2 (second column). For instance, "c6937_idle_median" would be the "median" value of the time series "idle" (the aggregation.cpu-average.percent.idle KPI) for the node whose identifier is "c6937". Thus, we obtain matrices similar to the one shown in Figure 4 per feature of each KPI and per node, instead of having the huge set of values included in the time series.

Once the features are extracted (per KPI and per node), we have normalized the results to be within a range of [0,1] to uniform the results, which are expressed using different measurement units. Table B (appendix) summarizes the variances obtained for all the scaled features and Figure 5 shows, for instance, the variance of the 36 features of one of the KPIs (Idle in Table 2). In this specific case, the feature percentage_of_reoccurring_values_to_all_values has the highest variance value.

**Table 4.** Extracted features (Tsfresh) and their parameters.

| Feature | Parameters | Feature | Parameters |
| --- | --- | --- | --- |
| length | None | longest_strike_below_mean | None |
| abs_energy | None | mean_change | None |
| mean | None | sample_entropy | None |
| median | None | standard_deviation | None |
| count_above_mean | None | percentage_of_reoccurring_values_to_all_values | None |
| count_below_mean | None | percentage_of_reoccurring_datapoints_to_all_datapoints | None |
| absolute_sum_of_changes | None | fft_aggregated | {Centroid, variance, skew, kurtosis} |
| mean_abs_change | None | friedrich_coefficients | {0,1,2,3} |
| mean_second_derivative_central | None | spkt_welch_density_coeff | {2,5,8} |
| maximum | None | index_mass_quantile | {10,20,30,40,50,60,70,80,90} % |
| minimum | None | ar_coefficient | {0,1,2,3,4} |
| Skewness | None | augmented_dickey_fuller | {Teststat, pvalue, usedlag} |
| Kurtosis | None | time_reversal_asymmetry_statistic | Lag {1,2,3} |
| first_location_of_maximum | None | c3 | Lag {1,2,3} |
| first_location_of_minimum | None | quantile | {10,20,30,40,50,60,70,80,90} % |
| binned_entropy | None | autocorrelation | Lag {1,2,3,4,5,6,7,8} |
| variance | None | number peaks | {5,10,15,20,30,35,40,50,100} |
| longest_strike_above_mean | None | linear_trend | {Pvalue, rvalue, intercept, slope, stderr} |

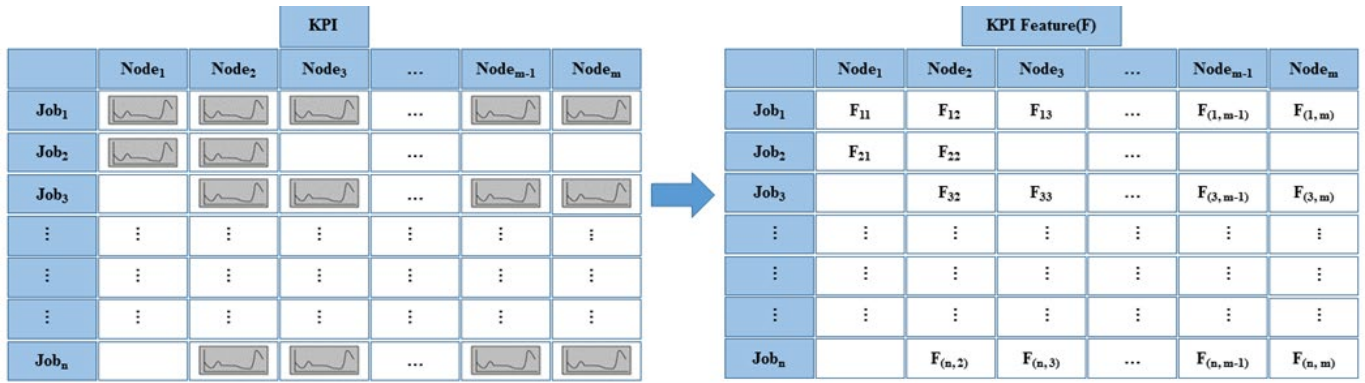

**FIGURE 4.** Dimensionality reduction based on feature extraction: on the left on KPI dataframe, on the right the reduced data.

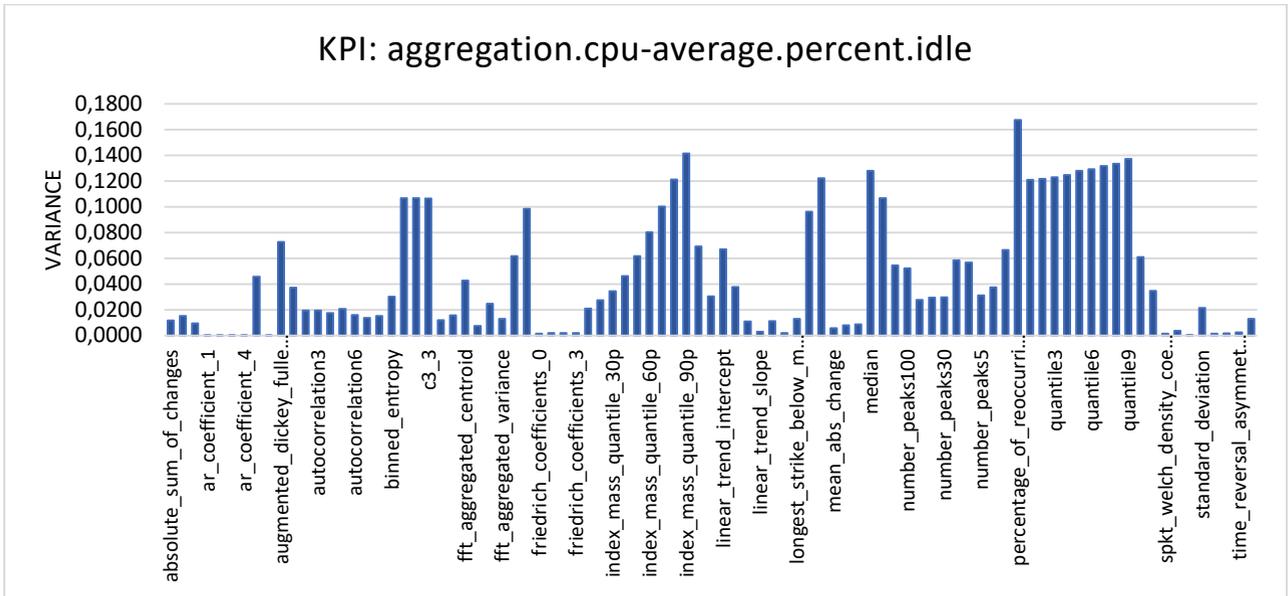

**FIGURE 5.** Variance of the 36 features extracted for one of the eleven KPIs.

### B. PHASE 2: SELECTION AND CLUSTERING

Although we have reduced the dimensionality after the first phase (feature extraction), in the second phase we go a step further with the aim to reduce even more the number of features, selecting only those ones that are relevant. With this aim, we apply two different approaches.

On the one hand, a literature-based feature selection was done applying the data in Table 1 (summary of the different approaches in the specialized literature) and the information in Table 4 (features obtained from the Tsfresh library). The intersection of both sources is summarized in Table 5: 29 features to be analyzed since they were considered as relevant for time series cluster analysis.

On the other hand, a variance-based feature selection was done, taking as relevant only those ones whose variance is greater than a previously defined threshold. Thus, as result of Phase 1, we have obtained a collection of 36 features that characterize the time series behavior of the different KPIs and nodes. These values were normalized to be within the range [0,1]. In this step, we try to reduce this number by using the variance threshold technique [79] [80], i.e. using a threshold to select only those features whose variance is equal or higher. By eliminating low variance features, we are trying to remove from the analysis those features that are not meaningful. We have used three different thresholds, 80%, 85%, and 90%, to perform this selection. Table 6 summarizes the results obtained. The first and second column represent the name of each one of the 11 KPIs. Then, the third, fourth, and fifth columns indicate, respectively, the results for the 80% threshold, 85% threshold, and 90% threshold: (i) the threshold value for the variance (second row); (ii) the number of features selected for all the KPIs (third row) and (iii) the subsequent rows indicate the name of the features selected per KPI.

**Table 5.** Literature-based selection of features.

| Features | Parameters | Features | Parameters |
|---|---|---|---|
| maximum | None | C3 (nonlinearity) | Lag {1, 2, 3} |
| minimum | None | time_reversal_asymmetry_statistic | Lag {1, 2, 3} |
| median | None | autocorrelation or Serial correlation | Lag {1, 3, 5, 8} |
| mean | None | sample entropy | None |
| index mass quantile | {10,20,30,40,50,60,70,80,90} % | linear trend | {Intercept, pvalue, rvalue, slope, stderr} |
| skewness | None | longest strike below mean | None |
| Kurtosis | None | variance | None |
| abs energy | None | longest strike above mean variance | None |
| absolute sum of changes | None | mean abs change | None |
| ar coefficient | K10  coeff(0, 1, 2, 3, 4) | mean second derivative central | None |
| augmented dickey fuller | {Usedlag, teststat, pvalue} | quantile | {10,20,30,40,50,60,70,80,90} % |
| binned entropy | Bins 10 | mean change | None |
| first location maximum | None | number of peaks | {5, 10, 20, 25, 50, 100} |
| first location minimum | None | percentage of reoccurring values to all values | None |
| percentage of reoccurring datapoints to all datapoints | None | | |

**Table 6.** Features variance thresholds and their corresponding selected features.

| Category | | Threshold percentage | 80% | 85% | 90% |
|---|---|---|---|---|---|
| | | Threshold | 0.16 | 0.12 | 0.09 |
| | | # of features selected from all KPIs | 3 | 7 | 21 |
| CPU usage | IDLE | | percentage_of_reoccurring_values_to_all_values | percentage_of_reoccurring_values_to_all_values | percentage_of_reoccurring_values_to_all_values |
| | | | | index_mass_quantile {90} % | index_mass_quantile {70,80,90} % |
| | | | | Median | median |
| | | | | Quantile{50,60,70,80,90} % | Quantile{10,20,30,40,50,60,70,80,90} % |
| | | | | | c3{1,2,3} |
| | | | | | first_location_of_minimum |
| | | | | | mean |
| | | | | | minimum |
| | | | | | maximum |
| | SYSTEM | | -- | first_location_of_minimum | first_location_of_minimum |
| | | | | | first_location_of_maximum |
| | WAIT | | -- | -- | index_mass_quantile {80,90} % |
| Network (interface) traffic | RX | | -- | -- | -- |
| | TX | | -- | -- | -- |
| IPMI | CPU1 | | -- | -- | -- |
| | CPU2 | | -- | -- | -- |
| | PW | | -- | -- | linear_trend {Pvalue} |
| | SYSTEM_TEMP | | -- | -- | None |
| System Load | SHORTTERM | | -- | -- | first_location_of_maximum |
| | | | | | minimum |
| | | | | | Number peaks{5} |
| Memory usage | MEMORY | | Augmented dickey full {pvalue} | Augmented dickey full {pvalue} | Augmented dickey full {pvalue} |
| | | | first_location_of_maximum | first_location_of_maximum | first_location_of_maximum |
| | | | | | maximum |
| | | | | | percentage_of_reoccurring_datapoints_to_all_datapoints |
| | | | | | percentage_of_reoccurring_values_to_all_values |

Finally, we apply clustering techniques within this phase following two different approaches. On the one hand, we apply the K-means algorithm to the jobs using as criteria each KPI individually, that is, using the 29 selected features (according to the literature-based criterion) as data to perform the clustering. On the other hand, we apply the K-Means algorithm to the jobs using as criteria the whole set of KPIs and using as data to perform the clustering the 3, 7 or 21 features in Table 6 for the variance-based criterion (80%, 85%, and 90% respectively). It is necessary to highlight that since we do not have a predetermined number of clusters (K), we propose to iterate this value from 2 to 30 to find out for which value of K the best clustering quality values are achieved. The results are summarized in Section VI.

### C. PHASE 3: VISUALIZATION

Visualization of the clustering results is not directly feasible because of the high-dimensionality of the jobs features used. Thus, we propose a methodology to visualize these results in a more appropriate way based on clustering only the most influential features, as Figure 6 shows. Therefore, we firstly obtain the three more influential features in clustering and, after that, we represent the clustering using only these three features.

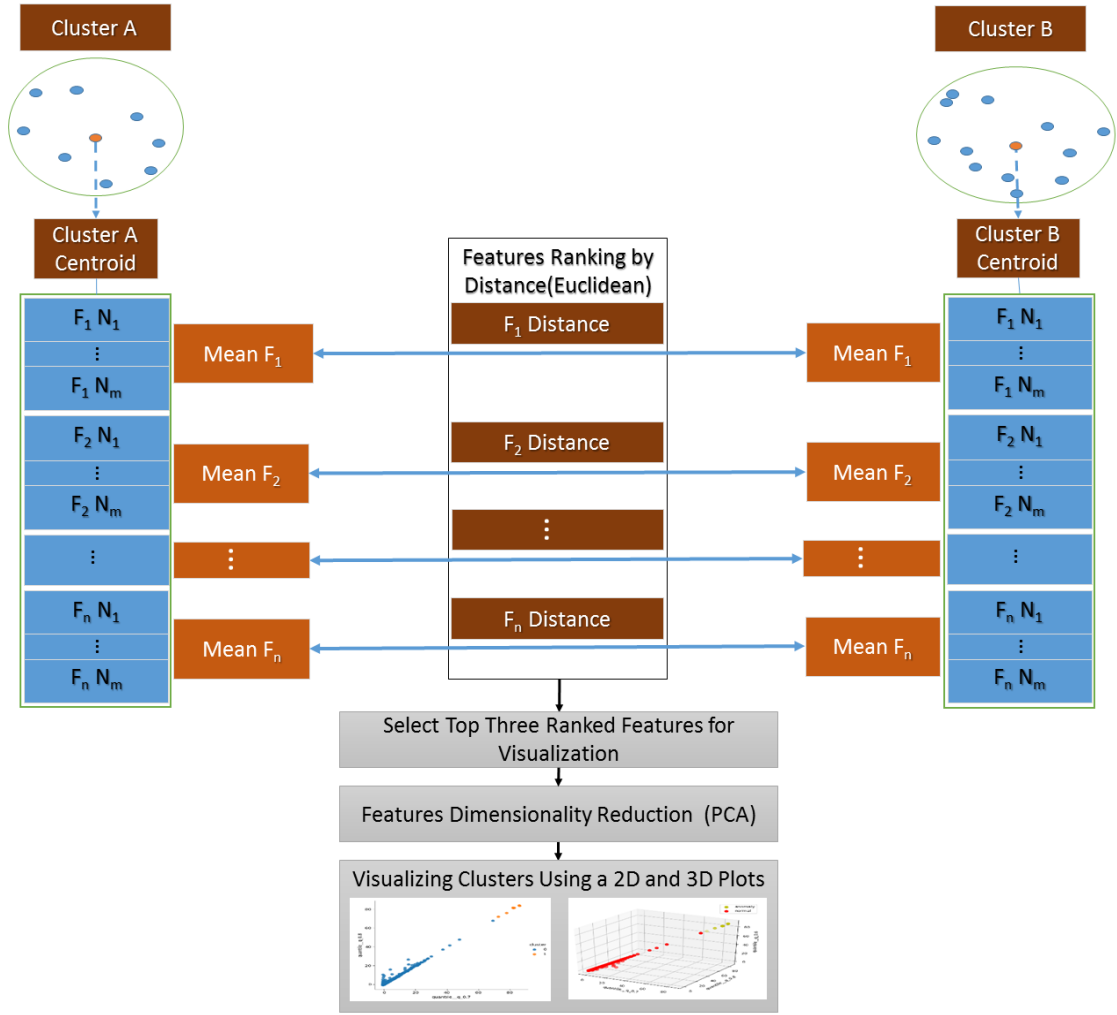

**FIGURE 6.** Selecting features for visualization.

The procedure is as follows. First, we obtain the centroids per cluster. Each centroid can be seen as a vector with as many components as computing nodes (n) multiplied by the number of features (m). Table 7 shows an example with two centroids for two clusters (0 and 1, in the first column).

Second, we calculate a vector with as many components as the number of features used for the clustering (m), whose component values are the mean of the feature values per node (mean of n values). After that, we calculate the Euclidean distance between the mean features of each vector. Later, we obtain the mean feature distances ranking in descending order to select the three top features. Table 8 shows an example with two clusters (i.e., two centroids) for all the features involved in the clustering at Phase 2.

**Table 7.** K-means centroids for the optimal number of clustering (2 clusters).

|   | c7339_system__first_location_of_minimum | c7343_system__first_location_of_minimum | ... | c6935_idle__quantile__q_0.7 | c6940_idle__median |
|---|---|---|---|---|---|
| 0 | 0.007707528 | 0.004937132 | ... | 0.227811218 | 0.202096027 |
| 1 | 0.416177977 | 0.045402307 | ... | 24.99583244 | 57.33320713 |

**Table 8.** K-means centroids for the optimal number of clustering (2 clusters) sorted after calculating the Euclidean distance.

|   | quantile__q_0.7 | median | first_location_of_maximum | ... | first_location_of_minimum |
|---|---|---|---|---|---|
| 0 | 0.272566967 | 0.257634946 | 0.01504391 | ... | 0.006514291 |
| 1 | 19.57893374 | 19.42098414 | 0.191517589 | ... | 0.087713726 |
| Distance | 19.30636677 | 19.1633492 | 0.176473679 | ... | 0.081199435 |

**Table 9.** The PCA results of the top three features with cluster labels.

| Job Id | quantile__q_0.7 | median | first_location_of_maximum | cluster |
|---|---|---|---|---|
| 3026217 | -0.565776 | -0.53378 | 3.921070 | 0 |
| 3033515 | -0.557738 | -0.52384 | 5.431736 | 0 |
| 3033516 | -0.513506 | -0.48461 | 3.455862 | 0 |
| ⋮ | ⋮ | ⋮ | ⋮ | ⋮ |
| 2894104 | -0.727134 | -0.691719 | -1.506159 | 1 |
| 2904125 | -0.756660 | -0.719323 | -0.994272 | 0 |
| 2894785 | -0.756660 | -0.719323 | -0.692337 | 0 |

Once we have identified the three top most influential features, these three features data were extracted separately from the original data set used in clustering to represent the clustering results in a 2D and 3D plot. Each feature matrix was 9,006 (job) x 195 (nodes), making the dimensionality of the three features data combined very high to be plotted. To have a more manageable volume of data for plotting, we have applied a 1-principal component using the PCA technique to each of the three feature data. After that, the three features PCA results were concatenated together with the cluster labels indexed by job identifier in only one dataframe. Table 9 shows an example of the jobs, where the three top ranked features PCA results are "quantile_q_0.7", "median" and "first_location_of_maximum" with their cluster labels. Finally, the first two features in the dataframe were used to plot the jobs clusters in a 2D plot, and the three features were used to plot the jobs clusters in a 3D plot.

## VI. EXPERIMENTAL RESULTS

We have used the data gathered from the CESGA HPC system (Section IV) to apply the methodology described in the previous section. Therefore, we have performed two experiments: on the one hand (experiment 1), clustering the jobs using a literature-based feature selection and, on the other hand (experiment 2), clustering the jobs using a variance-based feature selection. After that, we apply our methodology for visualization to depict the obtained results.

In experiment 1, we filtered the features of all the KPIs according to the information in Table 5, and then we performed as many K-means clustering as KPIs (eleven) separately. Thus, for instance, we clustered all the jobs in the data set using as criteria the set of features in Table 5 for the KPI IDLE, then we clustered all the jobs in the data set using as criteria the set of features in Table 5 for the KPI SYSTEM, and so on, for each one of the KPIs in Table 2. Since we do not know in advance the most appropriate number of clusters, we perform iterations from 2 to 30 to check the most suitable number of clusters (K) per KPI according to the Silhouette score.

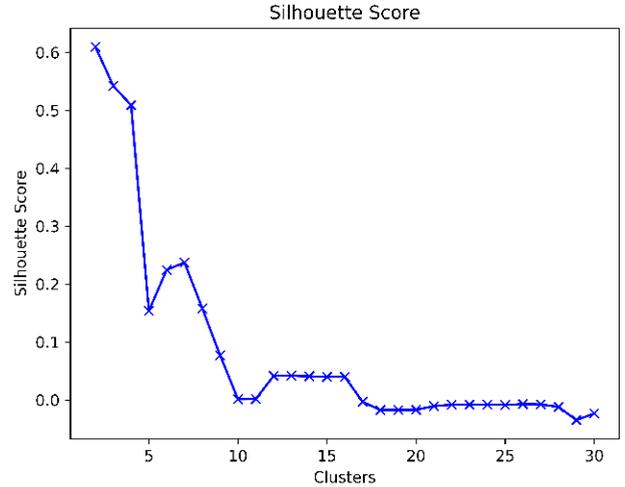

**FIGURE 7.** Silhouette scores of the KPI SYSTEM extracted features.

The obtained results are summarized in Figure 8 that shows the optimal number of clusters (K) and the quality score obtained per each one of the eleven KPIs. As a conclusion, we can state that the KPI SYSTEM (aggregation.cpu-average.percent.system) gives the best clustering result, followed by the KPI IDLE (aggregation.cpu-average.percent.idle) that obtains a very similar Silhouette score. Figure 7 shows the KPI SYSTEM Silhouette scores for each (K), which reveals the 2 clusters are the optimal number of clusters with Silhouette score of 0.6096.

In order to visualize these results, we applied our approach (Section V.C) to plot the clusters obtained by applying the top KPI clustering result (CPU SYSTEM, aggregation.cpu-average.percent.system) and its subset of features: a 9,006 x 13,065 matrix. Since the optimal number of clusters for this KPI CPU SYSTEM was 2, then the K-means centroids obtained can be represented as a 2 x 13,065 matrix. Afterwards, we grouped the centroids and calculated the mean value and the distance of each feature to identify the three most relevant features, in our case "index _mass_quantile__q_0.9", "quantile__q_0.9" and "quantile__q_0.8".

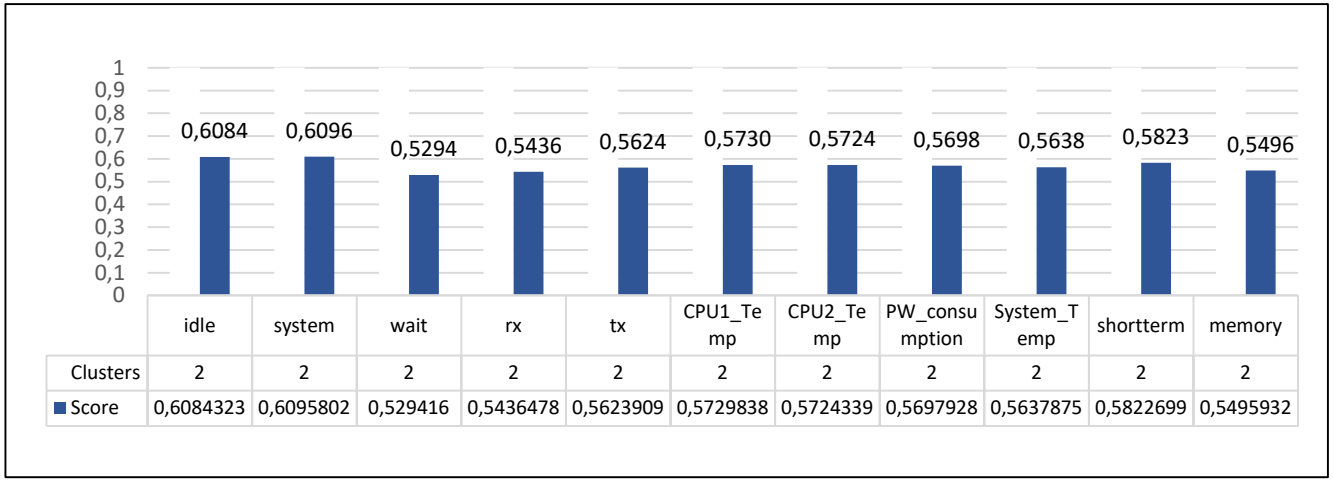

**FIGURE 8.** Silhouette score of clustering each jobs KPI extracted features individually.

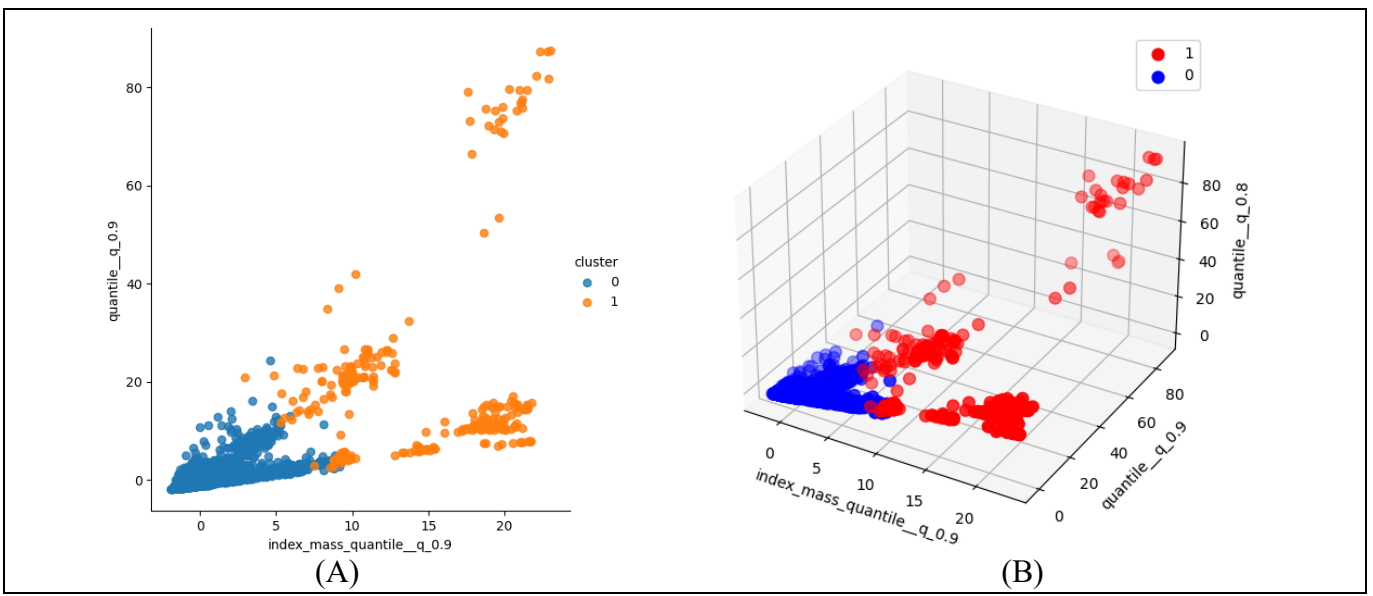

**FIGURE 9.** 2D and 3D clustering plot using the top three features of the CPU SYSTEM KPI.

Finally, each of these three features was dimensionally reduced to a 1-principal Component using PCA. Figure 9. (A) shows the 2D plot using the top two most significant features for the KPI CPU SYSTEM ("index _mass_quantile__q_0.9" and "quantile__q_0.9"), whereas Figure 9.(B) shows the 3D plot using the top three most significant features (adding "quantile__q_0.8"). The jobs in cluster 0 show higher cohesion than jobs in cluster 1 in both plots. Additionally, jobs in cluster 0 have lower values for the selected features than jobs in cluster 1. We can infer, from this behavior, that jobs in cluster 0 were executed more efficiently in the nodes of the HPC system, with less IO wait time, whereas jobs in cluster 1 group the jobs with the opposite tendency: more IO wait time and less efficient computation.

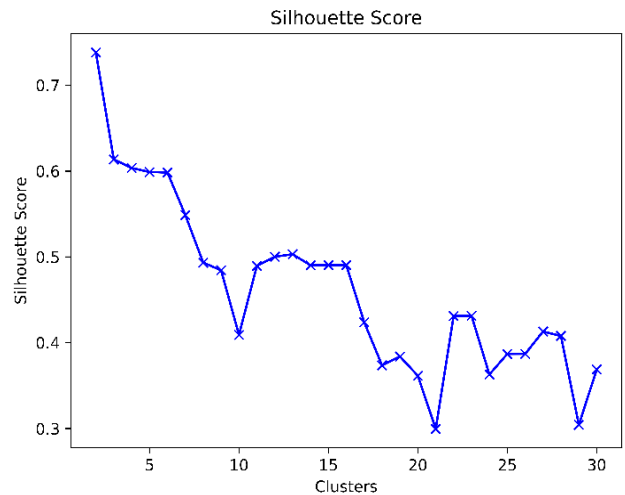

**FIGURE 10.** The Silhouette score for the 85% threshold features.

**Table 10.** K-means results for the (80%, 85%, 90%) threshold features.

| Variance threshold to select the most relevant features | Optimal K | Silhouette score |
|---|---|---|
| 80% Threshold | 2 | 0.6156 |
| 85% Threshold | 2 | 0.7382 |
| 90% Threshold | 2 | 0.5553 |

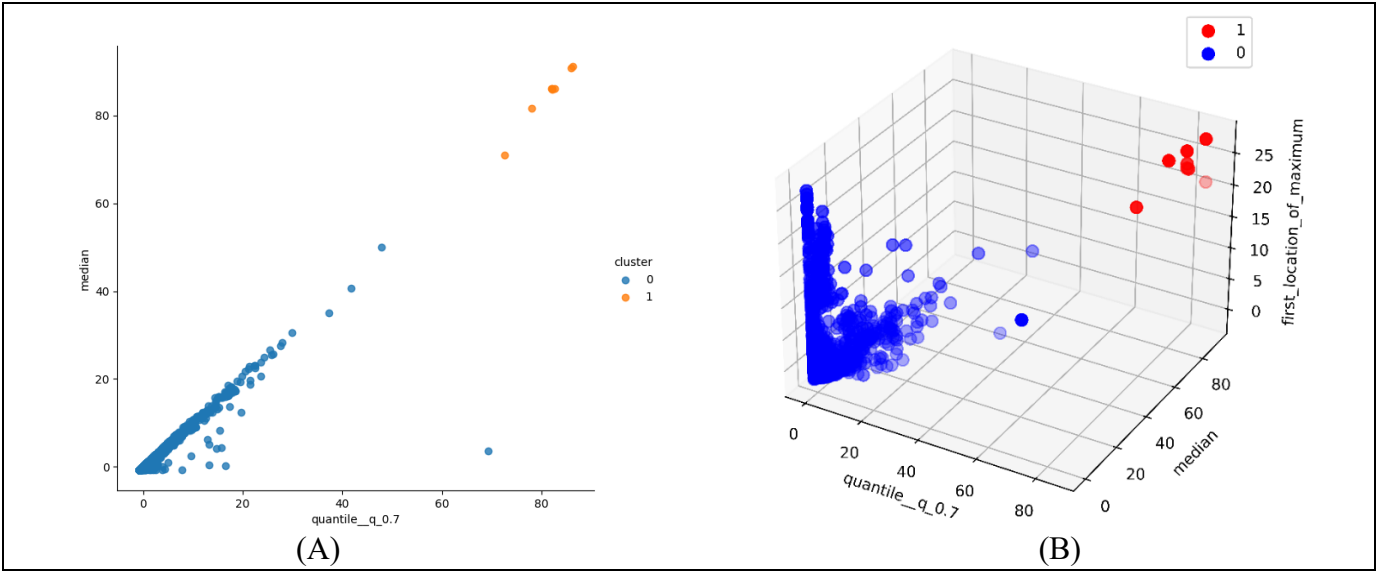

**FIGURE 11.** 2D and 3D clustering plot using the top three features in Experiment 2.

In experiment 2, we filtered the features of all the KPIs according to the information in Table 6 and using the three selected thresholds (80%, 85%, 90%). Then we performed a K-means clustering using jointly the data of the 11 KPIs, i.e. only one clustering procedure. Since we do not know in advance the most appropriate number of clusters, we performed iterations from 2 to 30 to check the most suitable number of clusters (K) per KPI according to the Silhouette score.

According to the obtained results, summarized in Table 10, the best Silhouette score was obtained for the 85% threshold with two optimal clusters, as shown in Figure 10. Thus, the next step is to plot the clusters for the 85% threshold, selecting the most relevant features of the 7 features that are involved in the clustering (see Table 6).

In this case, we have a 9,006 x 2,145 matrix that stores the information about the jobs. Since the optimal number of clusters is 2, then the K-means centroids obtained can be represented as a 2 x 2,145 matrix. After this, we selected the top three most relevant features involved in the clustering: "quantile__q_0.7", "median", "first_location_maximum". Later, the dimensionality of this data was reduced using PCA to obtain 1-principal component in order to have all the information to plot the clustering. Continuously, each of the three features components was concatenated together and used to plot the clusters in 2D and 3D.

Figure 11.(A) shows the 2D plot using the top two most significant features (quantile__q_0.7" and "median"), and Figure 11.(B) shows the 3D plot using the top three most significant features (adding "quantile__q_0.7"). We can conclude that jobs in cluster 0 have lower CPU idle time than jobs in cluster 1, which entails that the nodes workloads were correctly balanced during the jobs execution time. Besides, jobs on cluster 1 show an unusual behavior with higher "median" and "quantile__q_0.7" values. This might be a consequence of having higher IO tasks (network traffic rates), which has a negative correlation with CPU usage values. Analyzing also the 3D plot, we can infer that jobs in cluster 1 required the maximum memory usage at approximately the same time, whereas jobs in cluster 0 required from the maximum memory usage at different times during the global execution period.

## VII. DISCUSSION

As previously mentioned, we used the Silhouette score to evaluate the quality of the clustering in both approaches: when we applied the literature-based feature selection (one clustering per KPI) and when we applied the variance-based feature selection (one clustering using all the KPIs). In the first case, we obtained the quality results shown in Figure 8, where all the Silhouette scores are quite similar for all the KPIs. However, the IDLE KPI and the SYSTEM KPI provide better quality results with 2 clusters.

**Table 11.** Comparison of quality results to our previous approach in [7].

| KPIs | Experiment one clustering results | | Clustering results in [7] | |
| --- | --- | --- | --- | --- |
| | K | Silhouette Score | K | Silhouette Score |
| IDLE | 2 | 0.608 | 9 | 0.251 |
| SYSTEM | 2 | 0.609 | 4 | 0.212 |
| WAIT | 2 | 0.529 | 5 | 0.254 |
| RX | 2 | 0.543 | 8 | 0.390 |
| TX | 2 | 0.562 | 8 | 0.370 |
| CPU1 | 2 | 0.572 | 4 | 0.280 |
| CPU2 | 2 | 0.572 | 4 | 0.310 |
| PW | 2 | 0.569 | 4 | 0.290 |
| SYSTEM_TEMP | 2 | 0.563 | 4 | 0.340 |
| SHORTTERM | 2 | 0.582 | 6 | 0.088 |
| MEMORY | 2 | 0.549 | 5 | 0.130 |

We also compared these results to the ones we have obtained in our previous analysis [7], where PCA was used to face the data dimensionality problem. Table 11 summarizes the results of both approaches. According to this comparison, we can state that the methodology introduced in this paper provides better quality results.

In the second case, combining all the KPI data, the quality results are summarized in Table 10. We concluded that the 85% threshold is the most adequate to select the most relevant features. Finally, when comparing the two approaches: literature-based and variance-based selection, we can state that the best results are obtained if we considered the information given by all the 11 KPIs together for the clustering, instead of performing independent clustering per KPI.

The proposed feature-based clustering model can also be used to build a job anomaly prediction model. This, for sure, will be a useful tool for HPC DevOps engineers and technicians to early identify jobs/nodes whose behavior is not as expected. The application would trigger an alert if the KPIs of a job are performing similarly to the ones in the cluster of jobs with anomalies. Additionally, this tool would give to HPC DevOps engineers and technicians relevant information about the detected anomalies. Thus, which KPIs are the ones showing an unexpected behavior. This would contribute to identify the root of the problem in a timely manner and solve the issue as soon as possible. Consequently, this clustering model information would have a positive impact on minimizing the infrastructure cost.

## VIII. CONCLUSIONS

This paper introduces a methodology to cluster and visualize HPC jobs based on their performance KPIs. Our approach systematically identifies different job types (clusters) and supports their suitable visualization. Both clustering and visualization would help to manage and early detect performance problems in the nodes of the HPC system. We proposed two different approaches to deal with the high dimensionality problem inherent in these systems: a really high number of KPIs (44,280 in CESGA), a huge number of nodes (195 in CESGA), and a high sensing frequency in the HPC monitoring system (from 60 to 120 seconds). We focused our analysis on the following categories KPIs, CPU usage, Memory usage, IPMI, System Load, and Network (interface) traffic, which gave us an overwhelming amount of data: 35,761,290 samples on average per KPI and per job.

In order to reduce the high dimensionality of the data, we propose a methodology that faces the analysis of the collected time series in terms of its features, usually related to statistical behavior or global parameters (trend, seasonality, skewness, periodicity, etc.). The underlying idea is selecting only those features that are relevant for the jobs clustering, assuming that these features are usually correlated and redundant. We proposed to deal with this problem in a twofold approach: on the one hand, selecting the features based on the information gathered from the specialized literature and, on the other hand, selecting the features based on a threshold for their variance (the higher its variance, the more relevant). After performing our analysis, we concluded that the best approach is the second one using a threshold of 85% for the variance-based feature selection and combining the information provided by the 11 KPIs under study to perform the clustering. This approach gives the best clustering cohesion and separation. Thus, the results show that the dimensionality reduction techniques used in this study and our previous PCA-based study [7] enable a suitable way to cluster the jobs and show the convenience of using the KPIs related to the CPU usage (IDLE, SYSTEM) as the most suitable for clustering the HPC jobs.

Complementary to this clustering methodology, we have also defined a visualization procedure for the obtained clusters. First, we rank the features selected for the clustering according to their variance, with the aim of selecting the two and three top ones. After that, these two and three top features can easily visualize the obtained clusters in a 2D and 3D plot respectively.

We are currently planning to add more features and clustering algorithms to our current methodology, with the aim of improving the clustering results of HPC jobs and taking into consideration the computational power needed to execute such analysis. In addition, we will focus on the behaviors of the nodes in executing different jobs types, which will help us in building a forecasting model for node behavior.

# APPENDIX

**Table A.** Tsfresh extracted features definitions.

| # | Feature | Definitions of time series(x) extracted features |
|---|---|---|
| 1 | length | Number of samples of time series(x) |
| 2 | abs_energy | Interpreting the time series as the velocity of a particle with unit mass 2 |
| 3 | mean | Arithmetic mean of time series(x) |
| 4 | median | The middle of the sorted time series(x) values. |
| 5 | count_above_mean | Returns the number of values in x that are higher than the mean of x |
| 6 | count_below_mean | Returns the number of values in x that are lower than the mean of x |
| 7 | absolute_sum_of_changes | Returns the sum over the absolute value of consecutive changes in the series x. |
| 8 | mean_abs_change | Arithmetic mean of absolute differences between subsequent time series(x) values |
| 9 | mean_second_derivative_central | Returns the mean value of a central approximation of the second derivative. |
| 10 | maximum | Sample maximum of time series(x) |
| 11 | minimum | Sample minimum of time series(x) |
| 12 | skewness | Sample skewness calculated with adjusted Fisher-Pearson standardized moment coefficient |
| 13 | Kurtosis | Fourth central moment of time series (x) divided by the square of its variance |
| 14 | first_location_of_maximum | Returns the first location of the maximum value of x. The position is calculated to the length of x. |
| 15 | first_location_of_minimum | Returns the first location of the minimal value of x. The position is calculated to the length of x. |
| 16 | binned_entropy | This feature calculator bins the values of the time series sample into m equidistant bins. |
| 17 | variance | Expectation of the squared deviation of time series from its mean without bias correction |
| 18 | longest_strike_above_mean | Returns the length of the longest consecutive subsequence in x that is bigger than the mean of x. |
| 19 | longest_strike_below_mean | Returns the length of the longest consecutive subsequence in x that is smaller than the mean of x. |
| 20 | mean_change | Returns the mean over the differences between subsequent time series(x) values |
| 21 | sample_entropy | sample entropy of time series(x) |
| 22 | standard_deviation | Standard deviation of time series(x). |
| 23 | percentage_of_reoccurring_values_to_all_values | Returns the percentage of values that are present in the time series more than once. |
| 24 | percentage_of_reoccurring_datapoints_to_all_datapoints | Returns the percentage of non-unique data points. |
| 25 | fft_aggregated | Returns the spectral mean of the absolute Fourier transform spectrum. Returns the spectral variance of the absolute Fourier transform spectrum. Returns the spectral skew of the absolute Fourier transform spectrum. Returns the spectral kurtosis of the absolute Fourier transform spectrum. |
| 26 | friedrich_coefficients | Coefficients of polynomial h(x), which has been fitted to the deterministic dynamics of Langevin model |
| 27 | spkt_welch_density_coeff | Returns the estimates the cross power spectral density of the time series x at different frequencies. |
| 28 | index_mass_quantile | the relative index i where q% of the mass of the time series x lie left of i. |
| 29 | ar_coefficient | This feature calculator fits the unconditional maximum likelihood of an autoregressive AR(k) process |
| 30 | augmented_dickey_fuller | The Augmented Dickey-Fuller test checks the hypothesis that a unit root is present in a time series(x) sample |
| 31 | time_reversal_asymmetry_statistic | The theoretical symmetry of physical laws under the transformation of time reversal |
| 32 | c3 | Time series(x) non-linearity measure using a lag operator |
| 33 | quantile | Calculates the q quantile of x. This is the value of x greater than q of the ordered values from x. |
| 34 | autocorrelation | Calculates the autocorrelation of the time series with its lagged version |
| 35 | number peaks | Calculates the number of peaks of at least support n in the time series. |
| 36 | linear_trend | Calculate a linear least-squares regression for the values of the time series versus the sequence from 0 to length of the time series minus one |

**Table B.** KPIs scaled features variances.

| # | Features | system | wait | idle | CPU1 | CPU2 | PW | system temp | rx | tx | shortterm | memory |
|---|---|---|---|---|---|---|---|---|---|---|---|---|
| 1 | length | 0.0309 | 0.0309 | 0.0304 | 0.0330 | 0.0330 | 0.0330 | 0.0330 | 0.0334 | 0.0334 | 0.0335 | 0.0334 |
| 2 | abs_energy | 0.0119 | 0.0006 | 0.0152 | 0.0233 | 0.0276 | 0.0229 | 0.0175 | 0.0012 | 0.0027 | 0.0267 | 0.0081 |
| 3 | mean | 0.0346 | 0.0026 | 0.1222 | 0.0227 | 0.0231 | 0.0680 | 0.0244 | 0.0122 | 0.0215 | 0.0243 | 0.0865 |
| 4 | median | 0.0370 | 0.0027 | 0.1279 | 0.0235 | 0.0242 | 0.0702 | 0.0237 | 0.0122 | 0.0215 | 0.0306 | 0.0835 |
| 5 | count_above_mean | 0.0318 | 0.0162 | 0.0120 | 0.0132 | 0.0160 | 0.0130 | 0.0136 | 0.0333 | 0.0307 | 0.0149 | 0.0113 |
| 6 | count_below_mean | 0.0160 | 0.0237 | 0.0158 | 0.0171 | 0.0153 | 0.0136 | 0.0159 | 0.0317 | 0.0328 | 0.0211 | 0.0131 |
| 7 | absolute_sum_of_changes | 0.0224 | 0.0005 | 0.0116 | 0.0164 | 0.0209 | 0.0250 | 0.0290 | 0.0018 | 0.0011 | 0.0127 | 0.0007 |
| 8 | mean_abs_change | 0.0183 | 0.0019 | 0.0057 | 0.0115 | 0.0138 | 0.0167 | 0.0249 | 0.0006 | 0.0002 | 0.0009 | 0.0005 |
| 9 | mean_second_derivative_central | 0.0006 | 0.0014 | 0.0087 | 0.0009 | 0.0006 | 0.0018 | 0.0010 | 0.0004 | 0.0011 | 0.0069 | 0.0006 |
| 10 | maximum | 0.0246 | 0.0084 | 0.0961 | 0.0192 | 0.0202 | 0.0481 | 0.0177 | 0.0122 | 0.0215 | 0.0125 | 0.0903 |
| 11 | minimum | 0.0610 | 0.0012 | 0.1068 | 0.0392 | 0.0297 | 0.0726 | 0.0234 | 0.0122 | 0.0215 | 0.0958 | 0.0857 |
| 12 | skewness | 0.0071 | 0.0229 | 0.0346 | 0.0119 | 0.0104 | 0.0062 | 0.0015 | 0.0184 | 0.0009 | 0.0199 | 0.0025 |
| 13 | Kurtosis | 0.0038 | 0.0573 | 0.0692 | 0.0072 | 0.0058 | 0.0022 | 0.0011 | 0.0250 | 0.0006 | 0.0340 | 0.0025 |
| 14 | first_location_of_maximum | 0.1259 | 0.0591 | 0.0618 | 0.0763 | 0.0788 | 0.0450 | 0.0747 | 0.0328 | 0.0328 | 0.0949 | 0.1681 |
| 15 | first_location_of_minimum | 0.1315 | 0.0126 | 0.0983 | 0.0434 | 0.0439 | 0.0250 | 0.0420 | 0.0000 | 0.0000 | 0.0392 | 0.0524 |
| 16 | binned_entropy | 0.0681 | 0.0240 | 0.0301 | 0.0497 | 0.0461 | 0.0486 | 0.0232 | 0.0740 | 0.0201 | 0.0408 | 0.0833 |
| 17 | variance | 0.0063 | 0.0008 | 0.0128 | 0.0043 | 0.0045 | 0.0089 | 0.0047 | 0.0007 | 0.0004 | 0.0007 | 0.0011 |
| 18 | longest_strike_above_mean | 0.0030 | 0.0032 | 0.0019 | 0.0025 | 0.0048 | 0.0026 | 0.0042 | 0.0333 | 0.0307 | 0.0034 | 0.0167 |
| 19 | longest_strike_below_mean | 0.0024 | 0.0032 | 0.0131 | 0.0060 | 0.0047 | 0.0038 | 0.0121 | 0.0317 | 0.0328 | 0.0044 | 0.0090 |
| 20 | mean_change | 0.0017 | 0.0008 | 0.0078 | 0.0029 | 0.0027 | 0.0053 | 0.0019 | 0.0006 | 0.0002 | 0.0056 | 0.0005 |
| 21 | sample_entropy | 0.0676 | 0.0271 | 0.0609 | 0.0279 | 0.0250 | 0.0484 | 0.0252 | 0.0264 | 0.0091 | 0.0835 | 0.0631 |
| 22 | standard_deviation | 0.0123 | 0.0013 | 0.0216 | 0.0134 | 0.0135 | 0.0183 | 0.0099 | 0.0020 | 0.0012 | 0.0033 | 0.0041 |
| 23 | percentage_of_reoccurring_values_to_all_values | 0.0818 | 0.0452 | 0.1675 | 0.0076 | 0.0118 | 0.0074 | 0.0061 | 0.0520 | 0.0520 | 0.0256 | 0.1234 |
| 24 | percentage_of_reoccurring_datapoints_to_all_datapoints | 0.0563 | 0.0501 | 0.0664 | 0.0389 | 0.0424 | 0.0377 | 0.0389 | 0.0520 | 0.0520 | 0.0393 | 0.1008 |
| 25 | fft_aggregated_centroid | 0.0403 | 0.0450 | 0.0426 | 0.0278 | 0.0320 | 0.0373 | 0.0265 | 0.0081 | 0.0087 | 0.0300 | 0.0112 |
| | fft_aggregated_kurtosis | 0.0072 | 0.0586 | 0.0075 | 0.0062 | 0.0073 | 0.0230 | 0.0215 | 0.0060 | 0.0044 | 0.0063 | 0.0024 |
| | fft_aggregated_skew | 0.0160 | 0.0653 | 0.0247 | 0.0118 | 0.0134 | 0.0305 | 0.0412 | 0.0120 | 0.0273 | 0.0179 | 0.0050 |
| | fft_aggregated_variance | 0.0185 | 0.0135 | 0.0130 | 0.0133 | 0.0145 | 0.0215 | 0.0149 | 0.0068 | 0.0061 | 0.0125 | 0.0058 |
| 26 | friedrich_coefficients_0 | 0.0010 | 0.0042 | 0.0014 | 0.0127 | 0.0081 | 0.0045 | 0.0003 | 0.0003 | 0.0004 | 0.0018 | 0.0001 |
| | friedrich_coefficients_1 | 0.0005 | 0.0021 | 0.0019 | 0.0110 | 0.0077 | 0.0045 | 0.0009 | 0.0009 | 0.0002 | 0.0014 | 0.0002 |
| | friedrich_coefficients_2 | 0.0005 | 0.0021 | 0.0019 | 0.0102 | 0.0076 | 0.0044 | 0.0012 | 0.0012 | 0.0003 | 0.0014 | 0.0002 |
| | friedrich_coefficients_3 | 0.0006 | 0.0020 | 0.0019 | 0.0100 | 0.0078 | 0.0044 | 0.0006 | 0.0006 | 0.0003 | 0.0014 | 0.0003 |
| 27 | spkt_welch_density_coeff_2 | 0.0008 | 0.0003 | 0.0015 | 0.0022 | 0.0013 | 0.0016 | 0.0022 | 0.0005 | 0.0002 | 0.0017 | 0.0016 |
| | spkt_welch_density_coeff_5 | 0.0034 | 0.0004 | 0.0037 | 0.0028 | 0.0024 | 0.0031 | 0.0035 | 0.0002 | 0.0002 | 0.0043 | 0.0024 |
| | spkt_welch_density_coeff_8 | 0.0027 | 0.0003 | 0.0005 | 0.0011 | 0.0015 | 0.0022 | 0.0010 | 0.0003 | 0.0002 | 0.0006 | 0.0005 |
| 28 | index_mass_quantile_10p | 0.0053 | 0.0041 | 0.0211 | 0.0061 | 0.0062 | 0.0089 | 0.0103 | 0.0043 | 0.0105 | 0.0030 | 0.0095 |
| | index_mass_quantile_20p | 0.0059 | 0.0106 | 0.0276 | 0.0044 | 0.0044 | 0.0045 | 0.0055 | 0.0054 | 0.0131 | 0.0047 | 0.0032 |
| | index_mass_quantile_30p | 0.0068 | 0.0210 | 0.0345 | 0.0034 | 0.0034 | 0.0043 | 0.0063 | 0.0062 | 0.0144 | 0.0065 | 0.0037 |
| | index_mass_quantile_40p | 0.0058 | 0.0351 | 0.0461 | 0.0032 | 0.0032 | 0.0030 | 0.0062 | 0.0070 | 0.0155 | 0.0067 | 0.0038 |
| | index_mass_quantile_50p | 0.0068 | 0.0521 | 0.0618 | 0.0034 | 0.0033 | 0.0029 | 0.0065 | 0.0080 | 0.0168 | 0.0088 | 0.0034 |
| | index_mass_quantile_60p | 0.0074 | 0.0690 | 0.0802 | 0.0032 | 0.0029 | 0.0034 | 0.0048 | 0.0088 | 0.0172 | 0.0115 | 0.0029 |
| | index_mass_quantile_70p | 0.0084 | 0.0873 | 0.1003 | 0.0045 | 0.0029 | 0.0026 | 0.0087 | 0.0100 | 0.0176 | 0.0118 | 0.0024 |
| | index_mass_quantile_80p | 0.0080 | 0.1024 | 0.1213 | 0.0044 | 0.0038 | 0.0028 | 0.0080 | 0.0122 | 0.0185 | 0.0119 | 0.0019 |
| | index_mass_quantile_90p | 0.0062 | 0.1042 | 0.1415 | 0.0052 | 0.0056 | 0.0039 | 0.0103 | 0.0070 | 0.0131 | 0.0107 | 0.0010 |
| 29 | ar_coefficient_0 | 0.0054 | 0.0023 | 0.0094 | 0.0004 | 0.0002 | 0.0007 | 0.0013 | 0.0002 | 0.0002 | 0.0002 | 0.0002 |
| | ar_coefficient_1 | 0.0041 | 0.0007 | 0.0002 | 0.0144 | 0.0076 | 0.0097 | 0.0046 | 0.0005 | 0.0002 | 0.0133 | 0.0003 |
| | ar_coefficient_2 | 0.0046 | 0.0002 | 0.0002 | 0.0020 | 0.0020 | 0.0021 | 0.0044 | 0.0006 | 0.0003 | 0.0003 | 0.0003 |

|    | Feature | | | | | | | | | | | |
|----|---------|---|---|---|---|---|---|---|---|---|---|---|
|    | ar_coefficient_3 | 0.0005 | 0.0002 | 0.0002 | 0.0044 | 0.0005 | 0.0028 | 0.0020 | 0.0003 | 0.0007 | 0.0002 | 0.0002 |
|    | ar_coefficient_4 | 0.0005 | 0.0002 | 0.0002 | 0.0027 | 0.0002 | 0.0028 | 0.0009 | 0.0003 | 0.0005 | 0.0002 | 0.0003 |
| 30 | augmented_dickey_fuller_teststat | 0.0018 | 0.0002 | 0.0002 | 0.0028 | 0.0047 | 0.0041 | 0.0003 | 0.0011 | 0.0009 | 0.0049 | 0.0002 |
|    | augmented_dickey_fuller_pvalue | 0.0591 | 0.0192 | 0.0457 | 0.0379 | 0.0347 | 0.0179 | 0.0223 | 0.0518 | 0.0296 | 0.0384 | 0.1659 |
|    | augmented_dickey_fuller_usedlag | 0.0773 | 0.0714 | 0.0727 | 0.0652 | 0.0684 | 0.0395 | 0.0714 | 0.0733 | 0.0730 | 0.0529 | 0.0650 |
| 31 | time_reversal_asymmetry_statistic_lag_1 | 0.0013 | 0.0009 | 0.0014 | 0.0012 | 0.0012 | 0.0051 | 0.0016 | 0.0003 | 0.0003 | 0.0017 | 0.0004 |
|    | time_reversal_asymmetry_statistic_lag_2 | 0.0015 | 0.0006 | 0.0017 | 0.0012 | 0.0013 | 0.0039 | 0.0020 | 0.0003 | 0.0003 | 0.0009 | 0.0005 |
|    | time_reversal_asymmetry_statistic_lag_3 | 0.0019 | 0.0005 | 0.0024 | 0.0010 | 0.0012 | 0.0032 | 0.0019 | 0.0003 | 0.0003 | 0.0005 | 0.0006 |
| 32 | c3_1 | 0.0084 | 0.0024 | 0.1067 | 0.0107 | 0.0112 | 0.0656 | 0.0227 | 0.0035 | 0.0035 | 0.0013 | 0.0338 |
|    | c3_2 | 0.0085 | 0.0024 | 0.1066 | 0.0107 | 0.0113 | 0.0658 | 0.0227 | 0.0035 | 0.0035 | 0.0013 | 0.0338 |
|    | c3_3 | 0.0084 | 0.0024 | 0.1065 | 0.0107 | 0.0113 | 0.0659 | 0.0227 | 0.0035 | 0.0035 | 0.0013 | 0.0338 |
| 33 | quantile1 | 0.0253 | 0.0025 | 0.1210 | 0.0268 | 0.0265 | 0.0702 | 0.0232 | 0.0122 | 0.0215 | 0.0351 | 0.0857 |
|    | quantile2 | 0.0293 | 0.0027 | 0.1217 | 0.0250 | 0.0248 | 0.0666 | 0.0233 | 0.0122 | 0.0215 | 0.0334 | 0.0856 |
|    | quantile3 | 0.0313 | 0.0027 | 0.1230 | 0.0246 | 0.0252 | 0.0681 | 0.0234 | 0.0122 | 0.0215 | 0.0326 | 0.0829 |
|    | quantile4 | 0.0340 | 0.0027 | 0.1247 | 0.0240 | 0.0246 | 0.0685 | 0.0236 | 0.0122 | 0.0215 | 0.0317 | 0.0832 |
|    | quantile5 | 0.0370 | 0.0027 | 0.1279 | 0.0235 | 0.0242 | 0.0702 | 0.0237 | 0.0122 | 0.0215 | 0.0306 | 0.0835 |
|    | quantile6 | 0.0353 | 0.0027 | 0.1292 | 0.0239 | 0.0238 | 0.0732 | 0.0270 | 0.0122 | 0.0215 | 0.0143 | 0.0838 |
|    | quantile7 | 0.0359 | 0.0027 | 0.1317 | 0.0234 | 0.0234 | 0.0672 | 0.0271 | 0.0122 | 0.0215 | 0.0136 | 0.0842 |
|    | quantile8 | 0.0381 | 0.0027 | 0.1333 | 0.0231 | 0.0237 | 0.0605 | 0.0271 | 0.0122 | 0.0215 | 0.0136 | 0.0848 |
|    | quantile9 | 0.0410 | 0.0028 | 0.1371 | 0.0227 | 0.0234 | 0.0677 | 0.0240 | 0.0122 | 0.0215 | 0.0135 | 0.0855 |
| 34 | autocorrelation1 | 0.0490 | 0.0237 | 0.0371 | 0.0326 | 0.0317 | 0.0356 | 0.0332 | 0.0295 | 0.0027 | 0.0291 | 0.0220 |
|    | autocorrelation2 | 0.0397 | 0.0108 | 0.0195 | 0.0263 | 0.0302 | 0.0170 | 0.0309 | 0.0477 | 0.0066 | 0.0228 | 0.0250 |
|    | autocorrelation3 | 0.0374 | 0.0074 | 0.0194 | 0.0281 | 0.0293 | 0.0167 | 0.0297 | 0.0533 | 0.0092 | 0.0200 | 0.0391 |
|    | autocorrelation4 | 0.0314 | 0.0043 | 0.0175 | 0.0225 | 0.0273 | 0.0101 | 0.0243 | 0.0553 | 0.0116 | 0.0195 | 0.0326 |
|    | autocorrelation5 | 0.0260 | 0.0069 | 0.0206 | 0.0200 | 0.0280 | 0.0103 | 0.0234 | 0.0571 | 0.0147 | 0.0195 | 0.0475 |
|    | autocorrelation6 | 0.0245 | 0.0126 | 0.0158 | 0.0233 | 0.0253 | 0.0110 | 0.0248 | 0.0589 | 0.0180 | 0.0173 | 0.0443 |
|    | autocorrelation7 | 0.0213 | 0.0055 | 0.0136 | 0.0238 | 0.0262 | 0.0102 | 0.0196 | 0.0606 | 0.0215 | 0.0158 | 0.0408 |
|    | autocorrelation8 | 0.0215 | 0.0050 | 0.0152 | 0.0238 | 0.0172 | 0.0092 | 0.0195 | 0.0626 | 0.0252 | 0.0157 | 0.0416 |
| 35 | number_peaks10 | 0.0547 | 0.0492 | 0.0544 | 0.0534 | 0.0594 | 0.0385 | 0.0228 | 0.0000 | 0.0000 | 0.0827 | 0.0129 |
|    | number_peaks100 | 0.0304 | 0.0283 | 0.0521 | 0.0173 | 0.0198 | 0.0135 | 0.0168 | 0.0000 | 0.0000 | 0.0704 | 0.0163 |
|    | number_peaks15 | 0.0414 | 0.0261 | 0.0277 | 0.0522 | 0.0626 | 0.0341 | 0.0149 | 0.0000 | 0.0000 | 0.0838 | 0.0146 |
|    | number_peaks20 | 0.0277 | 0.0295 | 0.0293 | 0.0480 | 0.0445 | 0.0323 | 0.0105 | 0.0000 | 0.0000 | 0.0890 | 0.0136 |
|    | number_peaks30 | 0.0446 | 0.0548 | 0.0296 | 0.0356 | 0.0414 | 0.0232 | 0.0174 | 0.0000 | 0.0000 | 0.0811 | 0.0163 |
|    | number_peaks35 | 0.0498 | 0.0557 | 0.0584 | 0.0340 | 0.0348 | 0.0232 | 0.0173 | 0.0000 | 0.0000 | 0.0797 | 0.0129 |
|    | number_peaks40 | 0.0551 | 0.0563 | 0.0566 | 0.0292 | 0.0326 | 0.0210 | 0.0188 | 0.0000 | 0.0000 | 0.0881 | 0.0114 |
|    | number_peaks5 | 0.0517 | 0.0337 | 0.0313 | 0.0548 | 0.0645 | 0.0443 | 0.0255 | 0.0000 | 0.0000 | 0.0940 | 0.0129 |
|    | number_peaks50 | 0.0371 | 0.0432 | 0.0374 | 0.0238 | 0.0296 | 0.0225 | 0.0192 | 0.0000 | 0.0000 | 0.0772 | 0.0097 |
| 36 | linear_trend_intercept | 0.0224 | 0.0017 | 0.0670 | 0.0227 | 0.0231 | 0.0531 | 0.0242 | 0.0122 | 0.0215 | 0.0085 | 0.0818 |
|    | linear_trend_pvalue | 0.0786 | 0.0530 | 0.0378 | 0.0512 | 0.0502 | 0.0921 | 0.0535 | 0.0006 | 0.0004 | 0.0568 | 0.0046 |
|    | linear_trend_rvalue | 0.0144 | 0.0048 | 0.0109 | 0.0215 | 0.0227 | 0.0127 | 0.0295 | 0.0363 | 0.0105 | 0.0110 | 0.0763 |
|    | linear_trend_slope | 0.0025 | 0.0007 | 0.0030 | 0.0022 | 0.0024 | 0.0042 | 0.0017 | 0.0006 | 0.0002 | 0.0033 | 0.0005 |
|    | linear_trend_stderr | 0.0032 | 0.0009 | 0.0111 | 0.0050 | 0.0048 | 0.0106 | 0.0063 | 0.0005 | 0.0013 | 0.0107 | 0.0043 |

ACKNOWLEDGMENT

The authors would like to thank the European Regional Development Fund (ERDF) and the Galician Regional Government, under the agreement for funding the atlanTTIC Research Center for Information and Communication Technologies (atlanTTIC), and the Spanish Ministry of Economy and Competitiveness, under the National Science Program (TEC2017-84197-C4-2-R). The authors would also like to thank the Supercomputing Center of Galicia (CESGA) for their support and the resources for this research.

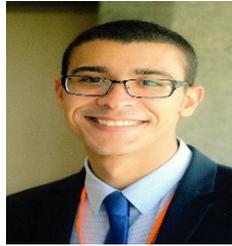
**MOHAMED S. HALAWA** is currently a PhD candidate in the Department of Telecommunication, University of Vigo, Spain. He received his master's degree in Computer Science from the Arab Academy for Science Technology and Maritime Transport, Cairo, Egypt, in 2015 and received his bachelor's degree in Business Information System from the Arab Academy for Science Technology and Maritime Transport, Cairo, Egypt, in 2009. His recent research focuses mainly on machine learning methods for anomaly detection.

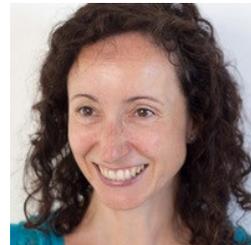
**REBECA P. DÍAZ REDONDO** is an Associate Professor at the Telematics Engineering Department at the University of Vigo and researcher in the Information & Computing Laboratory (atlanTTIC Research Center). She is currently working on defining appropriate architectures for distributed and collaborative data analysis, especially thought for IoT solutions, where computation must be on the edge of the network (Fog Computing). Rebeca has participated in more than 40 projects and 25 works of technological transfer through contracts with companies and/or public institutions. She is currently involved in the scientific and technical activities of several national and European research & educative projects.

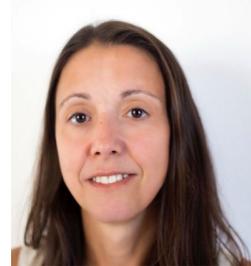
**Ana Fernández Vilas** is an Associate Professor at the Department of Telematics Engineering of the University of Vigo and researcher in the Information & Computing Laboratory (atlanTTIC Research Center). She received her PhD in Computer Science from the University of Vigo in 2002. Her research activity at I&C lab focuses on applied data science, specially, advanced machine learning for social Intelligence & IoT scenarios. She looks for solutions in the field of distributed deployment of data mining, training and model construction as a collaboration among nodes at edge, fog and cloud levels. Also, she is involved in several mobility & cooperation projects with North African countries & Western Balkans.